\documentclass[preprint,12pt]{elsarticle}

\usepackage{amssymb}
\usepackage{bm}
\usepackage{subfigure}
\usepackage{algorithm}
\usepackage{algorithmic}

\journal{arxiv.org}

\newcommand{\eqn}[2]
{
\begin{equation}\label{#1}
#2
\end{equation}
}

\newcommand{\eqna}[1]
{
\begin{eqnarray}
#1
\end{eqnarray}
}

\newtheorem{theorem}{Theorem}
\newtheorem{lemma}{Lemma}
\newproof{proof}{Proof}

\newcommand{\thm}[2]
{
\begin{theorem}\label{#1}
#2
\end{theorem}
}

\newcommand{\pf}[1]
{
\begin{proof}
#1
\end{proof}
}

\newcommand{\lem}[2]
{
\begin{lemma}\label{#1}
#2
\end{lemma}
 }

\newcommand{\fig}[4]
{
\begin{figure}[!htp]
\centering
\includegraphics[angle=0, width=#3\textwidth]{#4}
\caption{#2}\label{#1}
\end{figure}
}

\newcommand{\nn}
{
\nonumber
}

\begin{document}

\begin{frontmatter}



\title{Linear NDCG and Pair-wise Loss}


\author{Xiao-Bo Jin and Guang-Gang Geng$^{\ast}$\footnote{Guang-Gang Geng is the corresponding author.}}
\address{xbjin9801@gmail.com; gengguanggang@cnnic.cn}

\begin{abstract}
Linear NDCG is used for measuring the performance of the Web content quality assessment in ECML/PKDD Discovery Challenge 2010.
In this paper, we will prove that the DCG error equals a
new pair-wise loss.
\end{abstract}


\end{frontmatter}



\section{Linear NDCG}
In ECML Discovery Challenge 2010, the evaluation
measure is a variant of the NDCG ($NDCG^{\beta}$).  Given the sorted
ranking sequence $g$ and all ratings $\{r_i\}_{i = 1}^{|S|}$, the
discount function and NDCG are defined as ($r_i \in \{0,1,\dots, L -
1\}$):
\begin{equation}\label{eqn:dc_ndcg}
DCG_g^{\beta} = \sum_{i = 1}^{|S|}r_i(|S| - i)\textrm{ ,
}NDCG^{\beta} = \frac{1}{DCG_{\pi}^{\beta}}DCG_g^{\beta},
\end{equation}
where $DCG_{\pi}^{\beta}$ is the normalization factor that is DCG in
the ideal permutation $\pi$ ($DCG_g^{\beta} \le DCG_{\pi}^{\beta}$).
We call $\Delta DCG^{\beta} = DCG_{\pi}^{\beta} - DCG_g^{\beta}$ as
the DCG error. Specially, $DCG_{\pi}^{\beta} = mn + \frac{m(m -
1)}{2}$ for the bipartite ranking. It is worth noticing that the
above NDCG is different from the classical NDCG for the
query-dependent ranking, where the DCG function is (for the single
query):
\begin{equation}\label{eqn:delta_dcg_alpha}
DCG_g^{\alpha} = \sum_{i = 1}^{|S|}\frac{2^{r_i} -
1}{\log_2(i+1)}\textrm{ , }NDCG^{\alpha} =
\frac{1}{DCG_{\pi}^{\alpha}}DCG_g^{\alpha}.
\end{equation}

Consider the case of the query-dependent ranking with $L$ ratings.
For the given query, the dataset $S$ can be divided into $\{S_i\}_{i
= 0}^{L-1}$ according to the ratings of the instances. Generally, we
can define the empirical error for the multi-partite case:
\begin{equation}\label{eqn:multipartite_error}
\hat{R}(f) =  \frac{1}{Z} \sum_{0 \le a < b < L} \sum_{i =
1}^{|S_a|} \sum_{j = 1}^{|S_b|}(b - a)I[f(\bm{x}^{b}_i) <
f(\bm{x}^{a}_j)],
\end{equation}
where $Z = \sum_{0 \le a < b < L} |S_a||S_b|$. Specially, we also
define the following unnormalized empirical error: \eqn{}{R(f) =
\sum_{0 \le a < b < L} \sum_{i = 1}^{|S_a|} \sum_{j = 1}^{|S_b|}(b -
a)I[f(\bm{x}^{b}_i) < f(\bm{x}^{a}_j)].}

\section{$NDCG^{\beta}$ and Pair-wise Loss}
In this section, we will prove the following conclusion:
\eqn{}{\Delta DCG^{\beta} = R(f).}

\thm{thm:ecoc}{ For $L$-partite ranking problem, the unnormalized
empirical error can be divided into the following form:
\begin{equation}
R(f) =  \sum_{0 \le a < b < L} \sum_{i = 1}^{|S_a|} \sum_{j =
1}^{|S_b|}(b - a)I[f(\bm{x}^{b}_i) < f(\bm{x}^{a}_j)] =
\sum_{k = 0}^{L - 2}R_{k}(f),
\end{equation}
where
\begin{equation}
R_k(f) = \sum_{a = 0}^{k} \sum_{b = k + 1}^{L - 1} \sum_{i =
1}^{|S_a|} \sum_{j = 1}^{|S_b|} I[f(\bm{x}^{b}_i) <
f(\bm{x}^{a}_j)].
\end{equation}
}

\pf{ For the convenience of the description, we represent the
conclusion as follows: \eqna{G^{L}(f) & = & \sum_{k = 0}^{L -
2}R_k(f) \nn \\ & = &  \sum_{k = 0}^{L - 2} \sum_{a = 0}^{k} \sum_{b
= k + 1}^{L - 1} \sum_{i = 1}^{|S_a|} \sum_{j = 1}^{|S_b|}
I[f(\bm{x}^{b}_i) < f(\bm{x}^{a}_j)]  \nn \\ & = & R^L(f) }

Now we prove the conclusion $G^{n}(f) = R^{n}(f)$ with the
mathematical induction on the variable $n$. If $n = 2$, the
conclusion trivially holds. Assume that the equation is true for
$n$, then we will prove the conclusion for $n + 1$. We have
\begin{eqnarray}\label{eqn:g_x}
G^{n+1}(f)  & = & G^{n}(f) + \sum_{k = 0}^{n - 2}\sum_{a = 0}^{k}
\sum_{i = 1}^{|S_a|} \sum_{j = 1}^{|S_b|} I[f(\bm{x}^{n}_j) <
f(\bm{x}^{a}_i)] \nonumber
\\  && + \sum_{a = 0}^{n - 1}
\sum_{i =
1}^{|S_a|} \sum_{j = 1}^{|S_b|}I[f(\bm{x}^{n}_j) < f(\bm{x}^{a}_i)] \nonumber \\
& = & G^{n}(f) + \sum_{k = 0}^{n-1} \sum_{a = 0}^k \sum_{i =
1}^{|S_a|} \sum_{j =
1}^{|S_b|}I[f(\bm{x}^{n}_j) < f(\bm{x}^{a}_i)] \nonumber \\
\end{eqnarray}
and
\begin{equation}\label{eqn:r_x}
R^{n+1}(f) = R^{n}(f) + \sum_{a = 0}^{n - 1} (n - a) \sum_{i =
1}^{|S_a|} \sum_{j = 1}^{|S_b|}I[f(\bm{x}^{n}_j) <
f(\bm{x}^{a}_i)].
\end{equation}

Finally, we can prove by the mathematical induction that the second
item of the right side in  (\ref{eqn:g_x}) equals to the
corresponding item in (\ref{eqn:r_x}). We can see that for $n
= 1$ it is trivially hold.

It follows that $G^{L}(f) = R^{L}(f)$ for all natural number with $L
> 1$. }

\lem{prop:exchange}{ For the bipartite ranking problems, any sorted
ranking sequence from $S = \{S_{+},S_{-}\}$ can be obtained by
exchanging at most $k = \min\{|S_{+}|,|S_{-}|\}$ times from the
ideal ranking sequence.}

\pf{ Given that there are $r (r \le m)$ negative instances in the
first $m$ positions and $s (s \le n)$ positive instances in the
remain $n$ positions.

Now we prove $s = r$ indirectly through the apagoge. If $s \neq r$,
without loss of generality, we assume $r > s$. It is known that
there are $r - s$ negative instances in the first $m$ positions
after $s$ exchanges. The exchanges occur among $s$ negative
instances in the first $m$ positions and $s$ positive instances in
the remain $n$ positions. Then the fact that we will get $r - s + n$
negative instances is in contradiction to $n$ negative instances.
Finally, we can conclude that $r = s \le \min\{|S_{+}|,|S_{-}|\}$. }

Next, we will prove

\thm{thm:delta_dcg}{ For the bipartite ranking problem, DCG errors
with \ref{eqn:dc_ndcg} equals the unnormalized expected
losss $R(f)$:
\begin{equation}
\Delta DCG^{\beta} = R(f) = \sum_{i = 1}^m \sum_{j = 1}^n
I[f(\bm{x}^{+}_i) < f(\bm{x}^{-}_j)].
\end{equation}
}

\begin{figure}[t]
\centering
  \subfigure{
    \begin{minipage}[b]{0.1\textwidth}
      \centering
      \includegraphics[width= \textwidth]{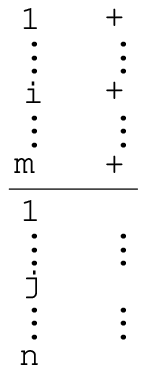}
    \end{minipage}}%
    \hspace{0.4in}
  \subfigure{
    \begin{minipage}[b]{0.13\textwidth}
      \centering
       \includegraphics[width= \textwidth]{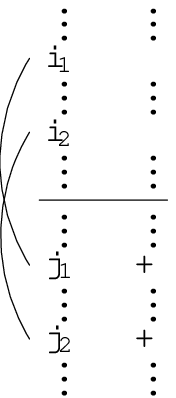}
    \end{minipage}}
  \caption{The ideal ranking sequence with its transformation. Left: the ideal ranking sequence,
  right: the ranking sequence with multiple exchanges}
  \label{fig:rank_exchange} 
\end{figure}

\pf{We know that any ranking sequence can be obtained by the
exchange operations from the ideal ranking sequence according to Prop.
\ref{prop:exchange}. Let $\{i_1,i_2,\cdots,i_k\}$$(1 \le i_1 <
i_2 < \cdots < i_k \le m)$ and $\{j_1,j_2,\cdots,j_k\}$$(1 \le j_1 <
j_2 < \cdots < j_k \le n)$ be the exchanged positions in the first
$m$ positions and the remain $n$ positions, respectively. As
depicted in Fig. \ref{fig:rank_exchange}, without loss of
generality, we exchange $i_{r}$ and $j_{r}$ for the r-th time.
First, we will compute the decrement relative to the ideal ranking
sequence for the r-th time
\begin{eqnarray}
\Delta_r DCG  & = & (m + n - i_r) - (m + n - (m + j_r)) \nonumber \\
           & = & m + j_r - i_r >= 1.
\end{eqnarray}
Now, we give a detailed explanation about the increment of the
unnormalize expected loss which is related to the position $i_r$ and
$j_r$. The increment due to the variation in the position $i_r$ will
be $m - i_r + r$ because there are $m - i_r$ positive instances in
the first $m$ positions and $r$ positive instances in the remain $n$
instances. As for the position $j_r$, the increment should be $j_r -
r$ since there are $j_r - 1 - (r - 1)$ negative instances in the
remain $n$ instances before $j_r$. In summary, we obtain the
increment $\Delta_r R(f) = m + j_r - i_r$. As a result, we conclude
that
\begin{equation}
\Delta DCG^{\beta} = \sum_{r = 1}^k \Delta_r DCG = \sum_{r =
1}^{k}\Delta_r R(f) = \Delta R(f).
\end{equation}
Notice that the initial value of $R(f)$ (the ideal ranking sequence)
is zero, this proves the theorem. }

\fig{fig:dcg_examples}{The example on the bipartite ranking shows
$\Delta DCG^{\beta} = R(f) = 4$,
   where $DCG_{\pi} = 12$ and $DCG_g = 8$.}{0.25}{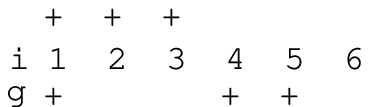}

Fig. \ref{fig:dcg_examples} gives an example to verify the
conclusion $\Delta DCG^{\beta} = R(f) = 4$.  The following theorem
shows that the conclusion $\Delta DCG^{\beta} = R(f)$ still holds
when extending to the multi-partite ranking problem.
\thm{thm:ndcg_dcg_error}{ For $L$-partite ranking problem, the DCG
errors with Eqn. (\ref{eqn:dc_ndcg}) equals $R(f)$:
\begin{equation}
\Delta DCG^{\beta} = \sum_{0 \le a < b < L} \sum_{i = 1}^{|S_a|}
\sum_{j = 1}^{|S_b|}(b - a)I[f(\bm{x}^{b}_i) < f(\bm{x}^{a}_j)).
\end{equation}
}

\pf{ From \ref{thm:ecoc}, we know that
\begin{equation}
R(f) = G(f) = \sum_{k = 0}^{L - 2}R_k(f).
\end{equation}

Then we will show that DCG in $L$-partite problem can be written as
the sum of the DCG measures of $L - 1$ bipartite problems. We divide
$DCG_{\beta}$ into
\begin{eqnarray}
DCG_{\beta}  & = & \sum_{i = 1}^{|S|}r_i(|S| - i) \nonumber \\
     & = & \sum_{i = 1}^{|S|} \sum_{k = 0}^{L - 2}I[k < r_i](|S| - i)
     \nonumber \\
     & = & \sum_{k = 0}^{L - 2}DCG_k,
\end{eqnarray}
where $DCG_k = \sum_{i = 1}^{|S|}I[k < r_i](|S| - i)$. For given
$k$, we can assign the instances with $r_i$ ($k < r_i$) to the
ranking $1$ and the others to the ranking $0$ to obtain a bipartite
ranking problem with the unnormalized empirical error \eqn{}{ R_k(f)
= \sum_{a = 0}^{k} \sum_{b = k + 1}^{L - 1} \sum_{i = 1}^{|S_a|}
\sum_{j = 1}^{|S_b|} I[f(\mathbf{x}^{b}_i) < f(\mathbf{x}^{a}_j)]. }

From \ref{thm:delta_dcg}, $\Delta DCG_k = R_k(f)$ holds. We
have $\Delta DCG = \sum_{k = 0}^{L - 2}\Delta DCG_k = \sum_{k =
0}^{L - 2}R_k(f) = R(f)$. }

\fig{fig:multi_dcg_examples}{The example on the multipartite ranking
shows $\Delta DCG_{\beta} = R(f) = 3$,
   where $DCG_{\pi}^{\beta} = 21$ and $DCG_g^{\beta} = 18$.}{0.25}{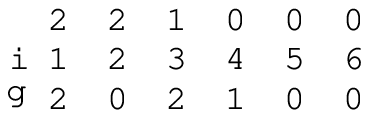}

The example in \ref{fig:multi_dcg_examples} supports our
conclusion about the DCG error and the unnormalized expected loss in
the multipartite ranking problem.


\end{document}